\documentclass[11pt,a4paper]{article}
\usepackage[hyperref]{emnlp2018}
\usepackage{times}
\usepackage{latexsym}
\usepackage[normalem]{ulem}
\usepackage{multirow}
\usepackage{amsmath}
\usepackage{lingmacros}
\usepackage{graphics}
\usepackage{tikz}
\usepackage{tikz-dependency}
\usepackage{url}
\usepackage{CJKutf8}
\usepackage{pgfplots}
\usepackage{pbox}
\usepackage{microtype}
\usepackage{todonotes}
\usepackage{enumitem}

\pgfplotsset{compat=newest}

\usepackage{subfig}

\usepackage{url}

\aclfinalcopy % Uncomment this line for the final submission

\newcommand\cue[1]{\textcolor{red}{\textbf{#1}}}

\title{Neural networks for cross-lingual negation scope detection}

\author{Federico Fancellu \ \ \ \ Adam Lopez \ \ \ \ Bonnie Webber\\
  ILCC, School of Informatics \\
  University of Edinburgh \\
  {\tt f.fancellu@ed.ac.uk}}

\date{}

\begin{document}
\maketitle
\begin{abstract}
Negation scope has been annotated in several English and Chinese corpora, and highly accurate models for this task in these languages have been learned from these annotations. Unfortunately, annotations are not available in other languages. Could a model that detects negation scope be applied to a language that it hasn’t been trained on? We develop neural models that learn from cross-lingual word embeddings or universal dependencies in English, and test them on Chinese, showing that they work surprisingly well. We find that modeling syntax is helpful even in monolingual settings and that cross-lingual word embeddings help relatively little, and we analyze cases that are still difficult for this task.
\end{abstract}

\section{Introduction}
Negation scope is the set of words whose meaning is affected by a word or morpheme expressing negation. For example, in (\ref{introex}), the words `you' and `drive' are in the scope of the negation cue `not'.

\enumsentence{\label{introex}
\uline{You} must \cue{not} \uline{drive} because it is dangerous.
}

Detecting negation scope is important for many applications, including biomedical information retrieval \citep[e.g.][]{morante2009metalearning}, sentiment analysis \citep[e.g.][]{councill2010what}, and machine translation \citep[e.g.][]{fancellu2014applying}. Its importance has prompted the development of several annotated corpora and classifiers that detect negation scope with high accuracy.

Most of this work is confined to English. Supervised machine learning systems require annotated data, and annotating data for negation scope requires substantial effort, both to adapt annotation guidelines to new languages \cite{altuna2017scope}, and for the annotation itself. As a consequence, there are only a handful of annotated datasets for languages other than English, such as the Chinese Negation and Speculation corpus \citep[CNeSp,][]{zou2016research}. We ask: \textit{Can we learn a model that detects negation scope in English and use it in a language where annotations are not available?}

\begin{CJK*}{UTF8}{gbsn}
\begin{figure*}[t]
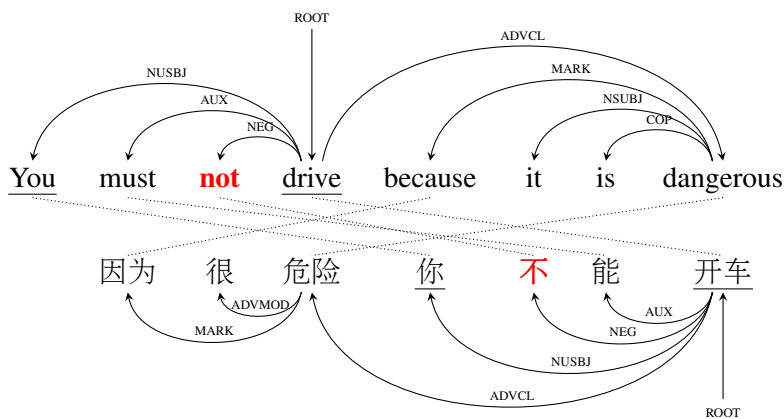

\centering
\begin{dependency}[theme = simple]
   \begin{deptext}[column sep=1em, row sep=4.5ex]
      \uline{You} \& must \& \cue{not} \& \uline{drive} \& because \& it \& is \& dangerous\\
          \& 因为 \& 很 \& 危险 \& \uline{你} \& \cue{不} \& 能 \& \uline{开车}\\
   \end{deptext}
   \draw[-,densely dotted] (\wordref{1}{5}.south)--(\wordref{2}{2}.north);
   \draw[-,densely dotted] (\wordref{1}{8}.south)--(\wordref{2}{4}.north);
   \draw[-,densely dotted] (\wordref{1}{1}.south)--(\wordref{2}{5}.north);
   \draw[-,densely dotted] (\wordref{1}{2}.south)--(\wordref{2}{7}.north);
   \draw[-,densely dotted] (\wordref{1}{3}.south)--(\wordref{2}{6}.north);
   \draw[-,densely dotted] (\wordref{1}{4}.south)--(\wordref{2}{8}.north);
   \deproot{4}{ROOT}
   \depedge{4}{1}{NUSBJ}
   \depedge{4}{2}{AUX}
   \depedge{4}{3}{NEG}
   \depedge{4}{8}{ADVCL}
   \depedge{8}{5}{MARK}
   \depedge{8}{6}{NSUBJ}
   \depedge{8}{7}{COP}
   
   \deproot[edge below]{8}{ROOT}
   \depedge[edge below]{8}{7}{AUX}
   \depedge[edge below]{8}{6}{NEG}
   \depedge[edge below]{8}{5}{NUSBJ}
   \depedge[edge below]{8}{4}{ADVCL}
   \depedge[edge below]{4}{3}{ADVMOD}
   \depedge[edge below]{4}{2}{MARK}
\end{dependency}
\caption{Example dependency parse for the sentence `You must not drive because it is dangerous' and its Chinese translation. While the word order differs in translation, each word in the negation scope stands in the same relation to the cue.}
\label{depex}
\end{figure*}
\end{CJK*}

To answer this question, we develop models on English using language agnostic features only and apply them to Chinese; though annotations are available for Chinese we use them only for testing to simulate our zero-resource setting. Our initial model is the state-of-the-art bidirectional LSTM (BiLSTM) of  \citet{fancellu2016neural}, initialized with cross-lingual word embeddings and universal part-of-speech (PoS) tags. But BiLSTMs are sensitive to word order, so we also experiment with a cross-lingual input representation that abstracts from word order---syntax in the form of universal dependencies \citep[UD,][]{demarneffe2014universal}---since we expect that for examples like that in Fig. \ref{depex}, this will give our model a more consistent view of the input across languages. To condition our model on UD syntax, we consider two different encodings: a Bidirectional DependencyLSTM \citep[D-LSTM below, modeled after the treeLSTM of][]{tai2015tree} and a Graph Convolutional Network \cite[GCN below,][]{marcheggiani2017encoding}

Our results show it is indeed possible to build models for cross-lingual negation scope detection with performance approaching that of a monolingual oracle. Modeling syntax in addition to surface word order is helpful, as shown by an ensemble of BiLSTM and D-LSTM models outperforming either model alone. Our results also show that cross-lingual word embeddings are not really necessary, suggesting that the model mainly relies on PoS, syntax, and punctuation boundaries---with the latter result reinforcing previous findings \citep{fancellu2017detecting}.

Finally, error analysis show that our best model performs better when the cue is in the same dependency substructure as its scope (as it is in Fig~\ref{depex}) and fails to capture phenomena related to negation scope, such as neg-raising, where lexical information is required.

\section{The task}
Our input is a sentence with a negation cue, which can be a word (e.g. `not') or a multi-word unit (e.g. `by no means') inherently expressing negation. Our task is to identify the set of words in the scope of the cue; we use gold cues and do not perform automatic cue detection. For example, \cue{\textbf{writing cues in bold red}} and \uline{underlining scopes}:

\enumsentence{\label{exinit}
\begin{enumerate}
\item[i] \uline{I} must \cue{not} \uline{go}
\item[ii] I do\cue{n't} think \uline{he should come} .
\item[iii] I did\cue{n't} miss the concert \uline{because I was sick} but because I was busy .
\end{enumerate}
}

Detecting negation scope is challenging because it often interacts with other semantic phenomena. To resolve (\ref{exinit}.i), the system needs to know that `must' scopes over negation but other modals (e.g. `should') do not. Likewise for neg-raising, as in (\ref{exinit}.ii), the presence of certain verbs like `think' or `believe' requires the negation scope to span the object clause (i.e. `I think he should not come'). Finally, in (\ref{exinit}.iii), the causal clause is in the scope despite the marker directly preceding the verb `miss'.

Similar interactions are attested in Chinese. However, the lack of markers for certain syntactic environments may pose different challenges, since scope boundaries are not defined explicitly. This is the case of clausal complements and descriptive clauses in the following examples which lack explicit markers (`to' and `that' in English).

\begin{CJK*}{UTF8}{gbsn}
\enumsentence{
\begin{enumerate}
\item[i.] \shortex{5}
{他 & 说 & \cue{不} & \uline{要} & \uline{等}}
{He & say & not & need to & go}
{``He says not to wait"}
\item[ii.] \shortex{6}
{我 & 有 & 衣服 & \cue{不} & \uline{要} & \uline{洗}}
{I & have & clothes & not & need & wash}
{``I have clothes that do not need to be washed"}
\end{enumerate}
}
\end{CJK*}

\section{Related work}\label{prevwork}
Automatically detecting negation scope at the string level has been tackled by a variety of classifiers \citep[e.g.][]{lapponi2012uio,packard2014simple} exclusively in English or Chinese monolingual settings  using language-specific heuristics or resources \citep[e.g. DeepBank][]{flickinger2012deepbank}. Corpora also reflect this limitation, with only one available in a language other than English. 

Recently, \citet{fancellu2016neural} proposed a BiLSTM model that can be easily repurposed to a new dataset without feature engineering, since it requires only word and universal PoS tags embeddings. Its performance is state-of-the-art in both English and Chinese, but to train it on another language we would still need annotations in that language. 

In the absence of annotated data in a target language, many work have underlined the usefulness of Universal Dependencies, a cross-lingually consistent syntactic annotation framework. \citet{tiedemann2015cross} and \citet{ammar2016many} explore the problem of parsing across the languages annotated for UD, while \citet{reddy2017universal} have converted UD annotation to logical form for universal semantic parsing and \citet{prazak2017cross} have used UD for cross-lingual SRL. 

\section{The models}

\subsection{BiLSTM}

Our BiLSTM model follows \citet{fancellu2016neural}, to which we refer the reader for further detail. Given a sentence $w = w_1$...$w_n$,  we encode $w_i$ as \textit{d}-dimensional embedding vector, {\bf w}$_i$ $\in {\rm I\!R^{d_w}}$. Alongside {\bf w}$_i$, we also encode information 1) about whether a word $w_i$ is a cue or not, encoded in a cue-embedding vector {\bf c}$_i$ $\in$ ${\rm I\!R^{d_c}}$ and 2) about the universal PoS tag of $w_i$, represented as a PoS embedding vector {\bf p}$_i$ $\in$ ${\rm I\!R{^d_p}}$. We then concatenate these vectors to yield the input {\bf x}$_i$ as follows:

\begin{figure}[h]
\begin{equation}
\begin{aligned}
{\bf x}_i=[{\bf w}_i;{\bf c}_i;{\bf p}_i]
\label{lstminput}
\end{aligned}
\end{equation}
\end{figure}

Our goal is to predict the negation scope $s \in \{1,0\}^{\mid w\mid}$, where $s_i=1$ if a token is part of the scope and 0 otherwise.

\subsection{Bidirectional Dependency LSTM (D-LSTM)}

We now turn to the encoding of a dependency tree, considering the example in Figure~\ref{depex}. We can traverse the tree bottom-up, from leaves to root, or top-down, from root to leaves. A top-down pass seems insufficient, since negation since cues are usually leaves as in the example. On the other hand, a bottom-up pass would fully encode the subtree rooted at the parent of the cue (in Fig.~\ref{depex}, `drive') but would not be able to encode information about the subordinate being out of scope. Hence we need a bi-directional model that can encode the tree bottom-up and top-down. But this is still insufficient unless the passes communicate: that is, if the bottom-up pass first collects information about the children of `drive', then the top-down pass can pick up that information and pass it downward, hence communicating information about `not' to its sibling nodes in scope.

The model accepts as input dependency trees. A dependency tree $g$ is a tuple (V,E), where $V_g$ is the set of word-nodes and $E_g$ the set of dependency edges. Each $e \in E$ is assigned a dependency label $l$. We define as $p(v)$ the parent of node $v$ and $C(v)$ the set of its children. $r$ is the root node.

We represent each word-node $v \in V$ as shown in Eq.\ref{initbtmup}. The input vector differs from the one used in the BiLSTM model in that we add an an extra embedding {\bf l} representing the dependency label of the word in $v$ and a linear transformation to allow multiple layers to be stacked together, as ${\bf x}_v$ can be replaced with the hidden state from a previous layer.

\begin{figure}[h]
\begin{equation}
\begin{aligned}
{\bf x}_v &= {\bf W}[{\bf x}_v;{\bf c}_v;{\bf p}_v;{\bf l}_v] + {\bf b}
\label{initbtmup}
\end{aligned}
\end{equation}
\end{figure}

The computation of the bottom-up pass is the same as in \citet{tai2015tree}. This pass returns the state $s_v^\uparrow$ = $\langle $ {\bf h}$_v^\uparrow$, {\bf c}$_v^\uparrow$ $\rangle$, where {\bf h}$_v^\uparrow$ and {\bf c}$_v^\uparrow$ are the hidden state and the memory cell of node $v$. 

To address the lack of bi-directionality in the original child-sum TreeLSTM of \citet{tai2015tree}, we add a second top-down pass where we feed the states computed during the bottom-up pass; in this our model is very similar to the one of \citet{chen2017improved}.

The top-down pass is similar to the bottom-up one but traverses the vertices in a topological order. To create a dependency between passes, we made the states computed during the bottom-up pass, ${\bf s_v^\uparrow}$, available in the form of additional weighted feature during the top-down pass.
We start by computing the representation of the root node $r$ as follows:

\begin{figure}[h]
\begin{equation}
\begin{aligned}
{\bf s}_r^\downarrow &=LSTM({\bf x}_r,s_r^\uparrow)
\end{aligned}
\end{equation}
\end{figure}

When computing the state of a node top-down, we use the parent state the same we did for the children states in the bottom-up pass. The hidden representation of the remaining nodes $v$, {\bf $s_v^\downarrow$}, is computed as follows:

\begin{figure}[t]
\centering
\includegraphics[scale=0.7]{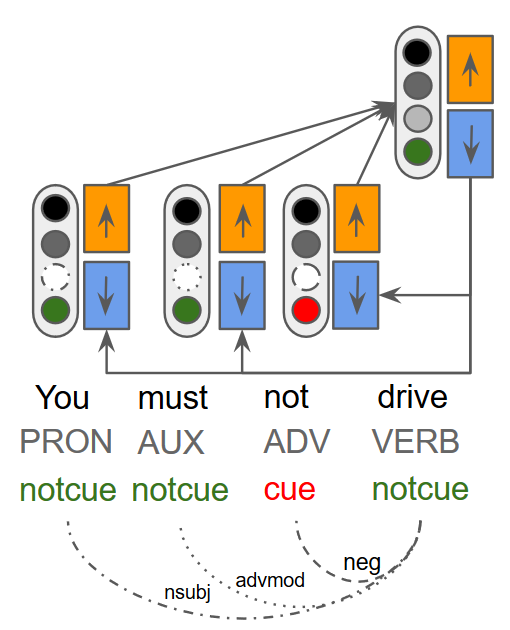}
\caption{The D-LSTM architecture. Each word is represented by the concatenation of word, universal PoS tags, dependency label and cue features. The latter is a binary feature which is 1 if the word is a cue (like `not') and 0 otherwise. The bottom-up pass builds from the leaves (`you', `must' and `not') to the root (`drive') and the top-down in the opposite direction. The states built during both passes are exemplified by the $\uparrow$ and the $\downarrow$ respectively.}
\label{treelstmfigure}
\end{figure}

\begin{equation}
\begin{aligned}
s_v^\downarrow &= LSTM({\bf x}_v, s_v^\uparrow, s_{p(v)}^\downarrow)
\end{aligned}
\end{equation}

After both passes are computed, we pass the hidden states obtained at the end of the top-down pass to the softmax layer to compute the probability of a given node to be inside or outside the scope of negation.\footnote{We also experimented with concatenating the two passes together but saw no difference in performance}

\begin{equation} \label{softmax}
\begin{aligned}
\hat p(y|{\bf h}_v) & = softmax({\bf W{\bf h}_v^\downarrow + b})
\end{aligned}
\end{equation}

A summary of the architecture is shown in Fig. \ref{treelstmfigure}.

\subsection{Graph convolutional networks} 

Our GCN is based on \cite{marcheggiani2017encoding}, to which we refer the reader for details. The intuition behind a GCN is that the hidden representation for each node in the tree is a function that aggregates information from its immediate neighbors. To communicate information between nodes that are not immediate neighbors, this process is iterated a fixed number of times, where each iteration corresponds to a neural network layer. GCNs do not assume that their input directed, so they have no notion of bottom-up or top-down traversal and do not distinguish between parent or child nodes; directionality is encoded explicitly into the neighborhood function.

The input to the model is a vector $[{\bf w_n;c_n;p_n}] \in {\rm I\!R}^{d^3}$, which is passed through a non-linearity or through a bi-LSTM before being fed to the GCN.

The computation for the hidden state of a given node $v$ takes into account: the hidden state of a neighbor node $n$; the directionality of the edge between $v$ and $n$ and the dependency label with its directionality specified. For each directionality a different weight matrix W$_{dir(u,v)}$ is used. Unlike the D-LSTM, information regarding the dependency label is not encoded in the input but in the bias vector b$^{l(u,v)}$.  This yields the following equation:
\begin{equation} \label{gcneq}
\begin{aligned}
{\bf h^{(K+1)}_v} &= ReLU(\\
& \sum_{u \in \mathcal{N}(v)} g^{(K)}_{v,u}({\bf W^{(K)}}_{dir(u,v)}{\bf h_v} + {\bf b^{l(u,v)}}))
\end{aligned}
\end{equation}
where g$_{(v,u)}$ is an edge-wise scalar gate to help weighing the importance of an edge-node pair amongst several neighbors and $K$ the current layer. However, whereas the original formulation of the GCN encodes information about the dependency labels in the bias term, we weight it alongside other input features. In this way, our GCN resembles the input of the D-LSTM. Our modification results in Eq~\ref{gcneq2}

\begin{small}
\begin{equation} \label{gcneq2}
\begin{aligned}
{\bf h^{(K+1)}_v} &= ReLU(\\
& \sum_{u \in \mathcal{N}(v)} g^{(K)}_{v,u}({\bf W^{(K)}}_{dir(u,v)}{\bf h_v} + {\bf W_l^{(K)} l}_{(u,v)}+ {\bf b}))
\end{aligned}
\end{equation}
\end{small}

\begin{figure}
\centering
\includegraphics[scale=0.7]{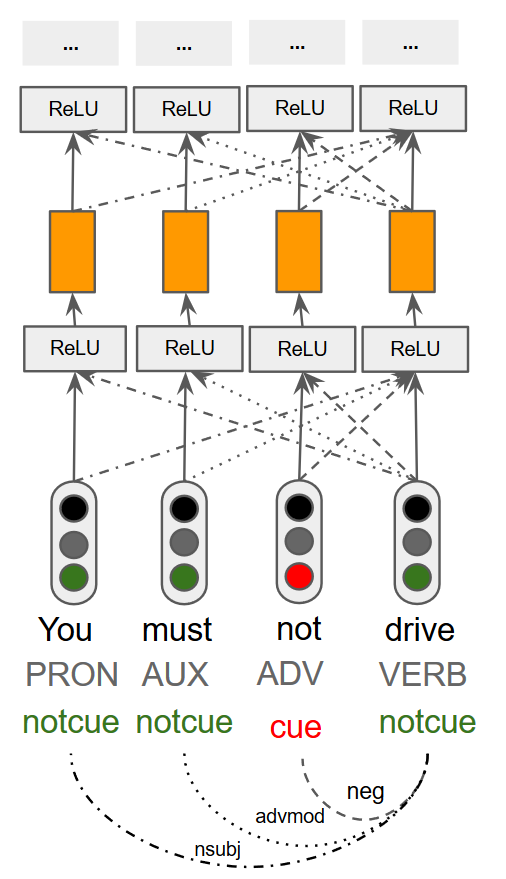}
\caption{The CGN architecture. Hidden representations are built by aggregating neighboring nodes in the dependency trees, as represented by the dashed lines. The node itself is also taken into consideration as shown by the straight lines. Information propagates by stacking up different layers.}
\label{gcnfigure}
\end{figure}

A summary of the architecture is shown in Fig. \ref{gcnfigure}.

\subsection{Ensemble}
Finally, we experiment with two different ensemble models, where we join together the BiLSTM with either the D-LSTM and GCN. We ensemble together our sequential classifier with each of the structured models, to see whether syntactic information can benefit from sequential information and viceversa. We experimented with three different ensemble techniques: a) jointly train the two systems and concatenate the output states of each word before softmax; b) feed the input through a BiLSTM layer (as shown in Eq.~\ref{lstminput}) before passing it through either the D-LSTM or the GCN (same to what \citet{marcheggiani2017encoding} have done to improve the performance of the GCN model) and c) voting. We found voting to achieve the best performance. We also experimented with different kind of voting and we opted for `confidence' voting, where for each word we choose the system where the absolute difference between probability of token being inside and outside the scope is larger. The results in the next section will be based on this last ensemble model.

\section{Data and experiment settings}
We experiment with \textsc{NegPar} \citep{liu2018negpar}, a parallel English-Chinese corpus of four Sherlock Holmes stories annotated for negation. Although the English side of \textsc{NegPar} leverages pre-existing annotations \citep[\textsc{ConanDoyleNeg}][]{morante2012conan}, most of it has been reannotated to better capture semantic phenomena related to negation scope like modality and 
neg-raising. Note that the Chinese translation often converts positive English statements to negative---for example, `This dress is cheap' becomes \begin{CJK*}{UTF8}{gbsn}这件衣服\cue{不}贵\end{CJK*} (`This dress is \cue{not} expensive'). Hence the Chinese contains more negation instances (Table~\ref{negpar_stats}).

\begin{table}[]
\begin{center}
\begin{tabular}{lcc}\hline\hline
& English & Chinese\\
\hline
train & 981 & 1206\\
dev & 174 & 230\\
test & 263 & 341\\ \hline
\end{tabular}
\caption{Number of negation instances in the train, dev and test set in the English and Chinese sides of \textsc{NegPar}}
\label{negpar_stats}
\end{center}
\end{table}

We obtain PoS tags and dependency parses using the Stanford Parser \cite{chen2014fast}. In preliminary experiments, we compared UD version 1 and version 2, which have an important difference: the negation-specific \textit{neg} relation in version 1 is replaced by the more general \textit{advmod} label in version 2. We observed that UD1 performs consistently better, so all experiments reported below are based on version 1. PoS tags are converted into universal PoS tags.\footnote{Mapping available at \url{https://github.com/slavpetrov/universal-pos-tags}} The word segmentation the Chinese side of  \textsc{NegPar} is based on also leverages Stanford toolkits \cite{chang2008optimizing}. When testing across language, we remove language-specific dependency tags (e.g.conj:and$\to$ conj).

\begin{table*}[]
\centering
\scalebox{0.85}{
\begin{tabular}{lcccccccccccc} \hline \hline
              & \multicolumn{4}{c}{English} & \multicolumn{4}{c}{English$\to$Chinese} & \multicolumn{4}{c}{Chinese} \\
              & P & R & F$_1$& PCS & P  & R  & F$_1$  & PCS & P & R & F$_1$ & PCS\\
              \hline
BiLSTM        & 85.29 & 89.76 & 87.47 & 55.89 & 69.43  & 70.45  & 69.94 & 18.64 & 77.71 & {\bf 79.35} & 78.52 & 33.14\\
D-LSTM        & 81.30 & 85.37 & 83.28 & 52.47 & 68.60  & 70.39  & 69.49 & 16.57 & 76.70 & 71.91 & 74.23 & 29.59\\
GCN           & 81.78 & 81.09 & 81.43 & 46.18 & 72.60  & 59.71  & 65.53 & 17.46 & 72.09 & 75.19 & 73.61 & 23.69\\

BiLSTM+D-LSTM & 87.86  & {\bf 89.77} & {\bf 88.80} & {\bf 61.98} & 72.03 & {\bf 72.89} & {\bf 72.46} & 21.01 & {\bf 81.47} & 77.89 & {\bf 79.64} & {\bf 40.53}\\
BiLSTM+GCN & {\bf 88.19} & 87.34& 87.77 & 59.54 & {\bf 74.62} & 69.24 & 71.92 & {\bf 23.65} & 78.02& 78.89 & 78.95 & 37.28 \\ \hline
\end{tabular}
}
\caption{(P)recision, (R)ecall, F$_1$, and percentage of correct Scope (PCS) for each model \textit{English}, where the model has been trained and tested in English; \textit{Chinese}, where the model has been trained and tested in Chinese; and \textit{English$\to$Chinese}, where the model has been trained in English and tested in Chinese.}
\label{ressummary}
\end{table*}

We experimented with three different cross-lingual word embeddings: a) embeddings pre-trained on Wikipedia data \footnote{Available at \url{https://github.com/Babylonpartners/fastText_multilingual}} where a linear transformation has mapped Chinese and English embeddings into a common space \citep{smith2017offline}; b) average cross-lingual word-embeddings \cite{guo2016representation}, where the embedding vector of a Chinese word is an average of the embedding vectors of its English translations and c) where we take as the embedding vector of a Chinese word the one of the English word with the highest translation probability. We found that c) consistently outperforms the other methods and that's what we are going to use in our experiments. We observed that method a) in particular suffers from a coverage problem since the embeddings cover only 64\% of the training vocabulary. We obtain translation probabilities from approximately 2 million sentences of the UN corpus \cite{rafalovitch2009united} using fast\_align \cite{dyer2013simple}.

Hyperparameter tuning was performed separately for each system. Both the D-LSTM and the GCN are optimized using Adam \citep{kingma2014adam}, with an initial learning rate of 0.005. We found 4 layers to yield the best performance for the GCN models. We use a dropout as regularizer; in the D-LSTM, dropout is performed on the output layer, whereas in the GCN we follow \citet{marcheggiani2017encoding} in performing dropout on the neighbors $N(v)$. 

We evaluate our models using precision, recall, and $F_1$ over the number of scope tokens; and using the percentage of \textit{full} scopes spans we correctly detect (PCS below). We evaluate our model cross-lingually by training in English and testing in Chinese (\textit{English$\to$Chinese}); and for comparison we test models that are trained and test monolingually, on only English or Chinese.

\section{Results and Discussion}

We summarize the results in Table \ref{ressummary} as follows:

\begin{enumerate}
\item {\bf Modeling syntax is useful, though not on its own}. The ensembles that incorporate syntax outperform other models on both F$_1$ and PCS in both the monolingual and cross-lingual settings, showing that syntax is indeed beneficial---note that they outpeform the state-of-the-art BiLSTM of \citet{fancellu2016neural,fancellu2017detecting}.\footnote{Our results are not directly comparable to those of \citet{fancellu2016neural,fancellu2017detecting} since the annotation of the English data is different.} The D-LSTM outperforms the GCN in the monolingual settings but the latter performs better in terms of full scope spans detected in the when training in English and testing in Chinese.
\item {\bf The BiLSTM model on its own outperforms either syntactic model on its own by a large margin}. Perhaps surprisingly, the BiLSTM performs on par with the D-LSTM in the cross-lingual setting as well, despite relying solely on surface word order. We investigate this in more detail below.
\item {\bf It is indeed possible to build a cross-lingual model of negation}, with performance that approaches that of a monolingual Chinese system.
\end{enumerate}

We also address the following questions:\\

\textit{Do all features contribute in the same way?} We perform feature ablation on our BiLSTM+D-LSTM ensemble  by either removing the cross-lingual word embedding feature (\textit{-w}) or the universal PoS embedding feature (\textit{-p}) from either or both model in the ensemble (Table \ref{ensemble_abl}). Results for the BiLSTM+GCN ensemble are similar.

Both ensembles show the same trend in that removing the cross-lingual word embedding or the universal PoS embedding feature from the structured models helps with both recall and $F_1$. This shows that both the D-LSTM and the GCN {\bf leverage the dependency structure as main feature for cross-lingual negation scope detection, with little impact from the other two features}. Results also show that for both ensembles, results are worse when removing the PoS embedding feature from the BiLSTM model, suggesting that the BiLSTM relies on PoS to model word order.

\begin{table}[h]
\centering
\begin{tabular}{cl|l|l|l|l}
  &  \multicolumn{4}{c}{~~~D-LSTM}       \\
\multicolumn{1}{l}{\multirow{2}{*}{}} & \multirow{2}{*}{}    & \multicolumn{1}{c}{all} & \multicolumn{1}{c}{-w} & \multicolumn{1}{c}{-p} & \multirow{2}{*}{} \\
\cline{2-6}\cline{2-6}
\multirow{9}{*}{BiLSTM}
& \multirow{3}{*}{all} & 75.09  & {\bf 78.27}  & 74.32  & P\\
& & 62.59 & 62.36  & 64.38 & R \\
&  & 68.27  & 69.55  & 69.17  & F$_1$ \\
\cline{2-6}
& \multirow{3}{*}{-w}  & 76.98 & 74.26 & 74.93  & P\\
& & 65.55  & {\bf 69.12} & 69.05  & R  \\
&  & 70.81   & 71.60  & {\bf 71.87}   & F$_1$           \\
\cline{2-6}
 & \multirow{3}{*}{-p}  & 75.38   & 73.2  & 71.86   & P  \\
&   & 56.68    & 56.62    & 59.79 & R   \\
&   & 64.71    & 64.04      & 65.28  & F$_1$
\end{tabular}
\bigskip
\caption{(P)recision, (R)ecall, and F$_1$ for the feature ablation experiments, where models trained on all the features (\textit{all}) are compared with models where cross-lingual word embeddings (\textit{-w}) and universal PoS tags (\textit{-p}) are removed. The row represents the three different ablated BiLSTM models, the columns the ablated D-LSTM models. Each cell represents an ensemble of these ablated models; for instance, the one with the highest F$_1$ (2$^{nd}$ row, 3$^{rd}$ column) is the ensemble of a BiLSTM where cross-lingual word embeddings are removed and a D-LSTM where universal PoS are removed.}
\label{ensemble_abl}
\end{table}

\textit{Why are BiLSTM useful in cross-lingual settings?} Results in Table 2 might be surprising considering that the \textit{sequential nature} of the BiLSTM does not adapt well with difference in word ordering that language can exhibit. This might be an artifact of the two languages used in the experiment, since English and Chinese have similar word order. However, this also could be explained by a striking observation by \citet{fancellu2017detecting}, who showed that recurrent classifiers are very accurate when negation scope is delimited by punctuation and sentence boundaries but inaccurate otherwise. For example, they would correctly predict the scope in Ex. (\ref{punct}.i), which we refer to as an \emph{easy} case, but not the one in Ex. (\ref{punct}.ii), a \emph{hard} case.

\enumsentence{\label{punct}
i. `\uline{She is} \cue{not} \uline{a princess}', said the queen .\\
ii. \uline{I} eat pizza but \uline{do} \cue{not} \uline{drink beer} .
}

To assess whether BiLSTM learns that punctuation is informative also in a cross-linguistic setting, we carry out two additional experiments.

First, we replicate the experiments \citet{fancellu2017detecting} and divide the development instances into two groups, the \textit{easy} instances, predictable by punctuation alone and the \textit{hard} instances where scope cannot be predicted by punctuation alone. If the predictions of the BiLSTM are guided by punctuation we would expect easy instances to be predicted correctly more often than hard ones. Results in Table \ref{punctanalysis} seems to confirm our prediction where the sequential model learns to use punctuation to detect negation scope. As for the hard cases we also noticed that in 47.6\% of the cases prediction begins or ends at a punctuation token.

\begin{table}[h]
\centering
\begin{tabular}{ccc} \hline\hline
condition & easy & hard\\
\hline
Chinese  & 48 & 22\\
English$\rightarrow$Chinese  & 27 & 5\\
\hline
\end{tabular}
\caption{PCS for \textit{easy} and \textit{hard} instances using the BiLSTM model on the development set.}
\label{punctanalysis}
\end{table}

Given these results, we expect performance to worsen when punctuation is removed from the training and test data, so that the model cannot rely on it. We tested this in a second experiment where we compare the performance of the BiLSTM model with the same model where punctuation in the input has been removed. Results in Table \ref{punctnopunct} confirm our hypothesis: the model with no punctuation performs significantly worse.

\begin{table}[]
\centering
\begin{tabular}{ccccc} \hline\hline
condition & P & R & F$_1$ & PCS\\
\hline
with punctuation & 66.2 & 71.0 & 68.5 & 13.8\\
without & 57.7 & 59.8 & 58.4 & 8.6\\
\hline
\end{tabular}
\caption{Comparison between two BiLSTM models in the cross-lingual task on the development set, one with (\textit{punct}) and one without (\textit{no punct.}) punctuation tokens.}
\label{punctnopunct}
\end{table}

\section{Error Analysis}

\textit{What is our model learning?} To analyze the performance of our best ensemble model, the BiLSTM+D-LSTM, we look at the \textit{syntactic environment} scope appears in. We approximate this by looking at the least common ancestor for all the nodes in the scope and by taking the label its parent edge; if the scope is discontinuous, we take into consideration the labels on top of all spans. For each of the most frequent dependency labels, we report token-level F$_1$, as well as the percentage of correct scope spans we recover (PCS). 

\begin{table}[]
\centering

\begin{tabular}{ccccc} \hline \hline
& \multicolumn{2}{c}{Chinese} & \multicolumn{2}{c}{English$\rightarrow$Chinese} \\
label & F$_1$ & PCS & F$_1$ & PCS\\
\hline
root & 76.4 & 41.1 & 70.5 & 22.9\\
conj & 81.6 & 32 & 81.7 & 25.8\\
ccomp & 77.5 & 43 & 66.5 & 10.7\\
nsubj  & 77.3 & 25 & 65.4 & 3.2\\
dep & 79.8 & 44 & 78.0 & 30.2\\
dobj & 70.8 & 6 & 63.2 & 9.3\\
nmod:prep & 68.9 & 0 & -- & -- \\
nmod & -- & -- & 55.9 & 5.2\\
advmod & 52.9 & 0 & 61 & 0\\ 
\hline
\end{tabular}
\caption{Analysis of the syntactic environment around the scope where the dependency label represents the parent of the least common ancestor of all the nodes in the scope. Labels are ordered from most to least frequent.}
\label{depanalysis1}
\end{table}

Results are shown in Table \ref{depanalysis1} for both English$\rightarrow$Chinese and Chinese settings. In the former, we notice that there is usually a substantial loss in performance in terms of PCS but not in terms of F$_1$, meaning that although the scope is not exactly captured the model is still able to correctly detect approximately the same proportion of tokens.

\begin{CJK*}{UTF8}{gbsn}\begin{figure*}
\enumsentence{\label{negraiseerr}
\begin{itemize}
\item[a.] \shortex{9}
{\{\uline{我} & 倒\} & \cue{没有} & \{想到 & \uline{你} & \uline{身上} & \uline{还} & \uline{有} & \uline{神经}\}}
{I & instead & have not & thing & you & on your body & still & have & strength}
{``I did not think you still had strength"}
\item[b.] \shortex{8}
{\{\uline{我}\} & \cue{并不} & \{认为 & \uline{我} & \uline{已} & \uline{弄清} & 全部 & \uline{情况}\}}
{I & not & believe & I & already & clarify & all & facts}
{``I really don't think I have clarified all the situation already"}
\end{itemize}
}
\enumsentence{\label{meiyouerr}
\shortex{11}
{要是 & \{\uline{我} & \cue{没}有 & \uline{弄错}\} & 的话 & 我们 & 的 & 当事人 & 已经 & 来 & 了}
{if & I & have not & mistake & if & we & DE & interested party & already & come & ASP}
{``If it wasn't for my mistake, our person of interest would have come already"}
}
\end{figure*}

In general, high performance is related to whether the cue is in the same dependency substructure as its scope. This happens when negation scope spans the entire sentence (`root') or when it spans complement (`ccomp') or coordinate clauses (`conj'), where the root is usually a verb which is the cue's parent. 

On the other hand, we found that the system do not detect full scope as well when the cue is in a different substructure than the scope. One of these cases involves neg-raising, which also requires lexical information and that the system always predict incorrectly. This is exemplified in (\ref{negraiseerr}.a) and (\ref{negraiseerr}.b) for the verbs 想到(`to think') and 认为(`to believe'), where the cue appears in the matrix clause but the scope spans the complement clause.

On the other hand, we found that in 8 out of 10 instances the universal quantifier was correctly predicted with respect to the scope of negation, as shown in the example below.

\enumsentence[]{\label{uniquanterr}
\shortex{8}
{\{\uline{礼靴} & \uline{和} & \uline{背心} & \uline{的} & \uline{钮扣}\} & 都 & \cue{没有} & \{\uline{扣好}\}}
{Boots & and & vest & DE & button & all & not & button up}
{``Boots and vest were both not buttoned up properly"}
}

Finally we found 8 cases where the systems does not distinguish the homographs 没有 (`have not'), where both characters are part of the cue and 没有 (`does not exist'), where only the first character is the cue and the second is the existential verb `there is' which is part of the scope. In these cases, the systems always include 没有 as part of the scope as shown in (\ref{meiyouerr}).

\end{CJK*}

\section{Conclusion}
Let us go back to our initial research question: when detecting negation scope in a language other than English, \textit{can we train a system to detect negation scope in English using language agnostic features and apply it to a language where no annotations are available?} Although not quite as accurate as an oracle monolingual model, we show that this is indeed possible by an ensemble of neural networks, where syntactic and surface word order complement each other. More interestingly, we show that the contribution of other cross-lingual features, such as bilingual word embeddings is minor compared to the information extracted from syntax. We also found that this applies to recurrent models as well, where structural information is extracted in the form of punctuation boundaries around a negated scope.

However, some phenomena related to negation scope, especially those requiring lexical information, are still missed by our system fail.

We also suggest that future work could apply this method to languages where negation is realized in divergent ways from that of English, like those displaying double and morphological negation.

\bibliographystyle{acl_natbib_nourl}
\bibliography{emnlp2018}

\end{document}